\documentclass[default,iicol]{sn-jnl}

\usepackage{amssymb}
\usepackage{color} 
\usepackage{amsmath}
\usepackage{algorithm}
\usepackage{placeins}
\usepackage{here}
\usepackage{multirow}
\usepackage{subfig}
\usepackage[strings]{underscore}
\usepackage{url}

\jyear{2022}%

\begin{document}

\title[Article Title]{\textsc{ConsInstancy:} Learning Instance Representations for Semi-Supervised Panoptic Segmentation of Concrete Aggregate Particles}

\author*[1]{\fnm{Max} \sur{Coenen}}\email{m.coenen@baustoff.uni-hannover.de}

\author[1]{\fnm{Tobias} \sur{Schack}}\email{t.schack@baustoff.uni-hannover.de}

\author[1]{\fnm{Dries} \sur{Beyer}}\email{d.beyer@baustoff.uni-hannover.de}

\author[2]{\fnm{Christian} \sur{Heipke}}\email{heipke@ipi.uni-hannover.de}

\author[1]{\fnm{Michael} \sur{Haist}}\email{haist@baustoff.uni-hannover.de}

\affil[1]{\orgdiv{Institute of Building Materials Science}, \orgname{Leibniz University Hannover}, \state{Germany}}

\affil[2]{\orgdiv{Institute of Photogrammetry and GeoInformation}, \orgname{Leibniz University Hannover}, \state{Germany}}


\abstract{We present a semi-supervised method for panoptic segmentation based on ConsInstancy regularisation, a novel strategy for semi-supervised learning. It leverages completely unlabelled data by enforcing consistency between predicted instance representations and semantic segmentations during training in order to improve the segmentation performance. To this end, we also propose new types of instance representations that can be predicted by one simple forward path through a fully convolutional network (FCN), delivering a convenient and simple-to-train framework for panoptic segmentation. 
More specifically, we propose the prediction of a three-dimensional instance orientation map as intermediate representation and two complementary distance transform maps as final representation, providing unique instance representations for a panoptic segmentation.
We test our method on two challenging data sets of both, hardened and fresh concrete, the latter being proposed by the authors in this paper demonstrating the effectiveness of our approach, outperforming the results achieved by state-of-the-art methods for semi-supervised segmentation.
In particular, we are able to show that by leveraging completely unlabeled data in our semi-supervised approach the achieved overall accuracy (OA) is increased by up to 5\% compared to an entirely supervised training using only labeled data. Furthermore, we exceed the OA achieved by state-of-the-art semi-supervised methods by up to 1.5\%.}

\keywords{ConsInstancy training, semi supervision, panoptic segmentation, instance representations, concrete aggregate}

\maketitle

\section{Introduction}
Today, concrete is the most dominant building material worldwide. 
Up to 80\% of the concrete's volume consists of fine and coarse aggregate particles (normally sizes of 0.1\,mm to 32\,mm) which are dispersed in a cement paste matrix.
The size distribution of the aggregates as well as the spatial distribution of the particles within the binder paste matrix are two criteria that substantially affect the quality characteristics of concrete. These include the concrete's stability and its workability in the fresh state, as well as the mechanical properties in the hardened state. 
The ability to automatically extract aggregate particles from visual data of concrete opens up new opportunities of large scale quality control, which is key in civil engineering to assess the quality of building components and to ensure the safety of building structures.
Towards this goal, we propose a CNN based method for the panoptic segmentation \cite{Kirillov2019} of concrete aggregate in images of both, hardened and fresh concrete. 

\begin{figure}[ht]
\begin{center}
		\includegraphics[width=1.0\columnwidth]{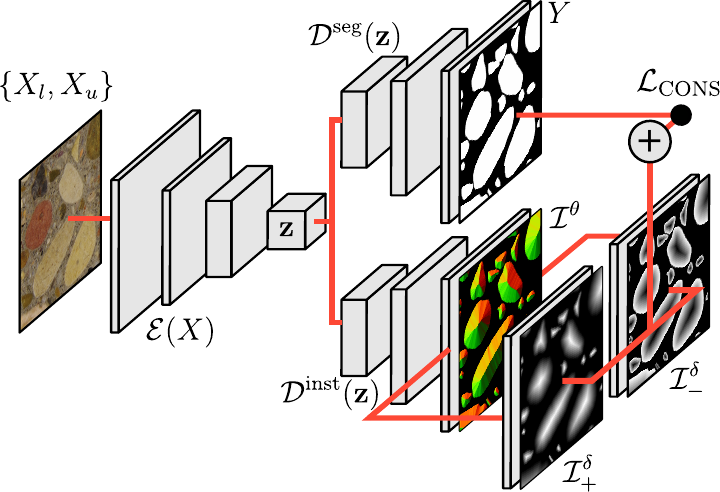}
	\caption{Overview of our framework for semi-supervised panoptic segmentation. Sharing a common encoder $\mathcal{E}$, a segmentation decoder $\mathcal{D}^\text{seg}$ is used to produce a semantic segmentation map and an instance decoder $\mathcal{D}^\text{inst}$ produces a three-dimensional orientation map $\mathcal{I}^\theta$ as intermediate, and two complementary distance transform maps $\mathcal{I}_+^\delta$ and $\mathcal{I}_-^\delta$ as final representation. The \textit{ConsInstancy} loss enforces consistency between the instance representations and the semantic segmentation map leveraging unlabelled data.}
\label{fig:overview}
\end{center}
\end{figure}

While a panoptic segmentation of images of hardened concrete delivers indications e.g.\ about the sedimentation stability of built components by considering the homogeneity of the particle distribution in the concrete, a panoptic segmentation of fresh concrete can be leveraged to derive workability characteristics and quality indicators of the material prior to its placement in the formwork. This, therefore, allows room for corrective or preventive measures, already during the construction process.
In \cite{Coenen2021a} and \cite{Wang2022} deep learning based approaches for the semantic segmentation of aggregate particles are proposed. While these approaches predict a semantic class (aggregate or cement paste matrix) for each pixel, we additionally determine a unique instance ID for each pixel, enabling the differentiation between individual particles, an extension which is especially relevant in the case of overlapping or neighbouring particles. 
However, a large amount of labelled training data for supervision is typically required for deep networks to learn the mapping of images to a semantic segmentation.
In the case of instance-aware segmentation of many, small, and potentially densely distributed objects, such as concrete particles, annotating large amounts of training data is immensely tedious. 
In this paper, we therefore propose a semi-supervised framework (cf.\ Fig.~\ref{fig:overview}) which leverages unlabelled data for the training process of a panoptic segmentation network in order to reduce the demand of annotated data and, thus, to improve performance. 
Our approach is applied to images of hardened concrete and, to the best of our knowledge, we are the first to propose the segmentation of aggregates also in images of fresh concrete. 

One successful line of work on semi-supervised semantic segmentation in the literature uses consensus regularisation during training by enforcing consistency between the predictions of a semantic segmentation of two or more decoder branches on unlabelled data \cite{Coenen2021a, Xiaomeng2021, Ouali2020, Peng2020, Zhang2020}. 
However, these approaches do not infer predictions at instance level. 
In \cite{Qizhu2018}, a weakly-supervised approach for panoptic segmentation is proposed, which leverages bounding box annotations in order to learn a segmentation at instance level.
However, the requirement of bounding boxes for training adds additional annotation effort to the learning procedure.
The question of how to best incorporate entirely unlabelled data to the learning of a \textbf{panoptic segmentation} is currently an open and active problem in research. 

Building upon the concept of consensus regularisation \cite{Chao2016, Ouali2020}, we make the following contributions in this paper:\\
\textbf{1)} With a 3D instance orientation map and two complementary semantic instance-aware distance transform representations, we propose novel instance representations that can be predicted in one path by a fully convolutional network (FCN), allowing the derivation of a panoptic segmentation of the input by one simple forward path.\\
\textbf{2)} In order to leverage unlabelled data to improve segmentation performance and to reduce the requirements of labelled data, we propose a \textit{ConsInstancy} regularisation, a novel semi-supervised training approach enforcing consistency between semantic instances and a semantic segmentation map predicted by a multi-task FCN. \\
\textbf{3)} We demonstrate on two data sets that our proposed method leads to superior results for the criteria of both, semantic and panoptic segmentation tasks, compared to state-of-the-art approaches. In this context, we propose our own and, to best of our knowledge, first data set of instance-wise annotated aggregate particles in images of fresh concrete, encouraging vision based research efforts towards precautionary, instead of retrospective, quality control of concrete \footnote{Source code is made publicly available here: \url{https://github.com/MaximilianCoenen/ConsInstancy.git}}.

\section{Related Work}
This section gives an overview of related work on instance representation and on current approaches for semi-supervised image segmentation.

\subsection{Instance representation}
A very common representation of instances in computer vision applications are \textit{bounding boxes} \cite{FasterRCNN}, typically represented as rectangular and axis aligned boxes, enclosing the instance.
In \cite{MaskRCNN}, bounding boxes are enriched by an \textit{instance mask}, delivering a pixelwise encoding of those pixels which are associated to the instance and, which are not. 
However, the approaches \cite{MaskRCNN, FasterRCNN} allow bounding boxes as well as the segmentation masks of different instances to overlap, rendering them unsuitable for the task of panoptic segmentation, in which it is necessary to assign every pixel to only one unique instance.

In \cite{CentroidNet2019}, instances are represented by two-dimensional vectors associated to each pixel and pointing to the nearest \textit{centroid} of an object.
While this representation allows to determine object centroids and consequently, enables locating and counting instances, it does not provide an instance-wise segmentation of the input.
To this end, the authors of \cite{CentroidNetV22021} enrich the centroid representation by an additional pixel-wise representation of vectors pointing to the nearest object boundary. 
Similarly, the authors of \cite{Xie2021} propose a \textit{polar mask} as instance representation, which defines each instance by the centre point of the object. In order to obtain the outline of the instance, a number of rays, sampled in uniformly distributed angular intervals, encode the distance to the closest boundary pixel along the respective ray. 
Likewise, in \cite{Schmidt2018, Weigert2020}, a \textit{star-convex polygon} is proposed to provide an instance representation. There, each pixel belonging to an object instance is allocated to the distances to the object's boundary along a set of predefined equidistant radial directions. 
In these representations, the instance boundaries are implicitly contained as polygonal shapes defined by  the radial rays and the associated length of each ray. 
Similarly, by proposing a \textit{complementary distance transform map} in this paper, we also employ an implicit encoding of the instance boundaries, however, providing a much simpler representation compared to the polar mask and the star-convex polygon.
In \cite{Bai2017}, the authors propose to predict a \textit{deep watershed transform}, which corresponds to a discretised distance transform map, for the task of instance segmentation.
Furthermore, as an intermediate representation, the authors make use of a two-dimensional \textit{direction map} in order to guide the learning process of the watershed transform.
In this map, each pixel is associated with a 2D unit vector pointing in the direction of the closest boundary point to that pixel.
However, since pixels belonging to non-object classes do not possess instance boundaries and, therefore, can not be assigned to a meaningful direction vector, this procedure requires a semantic segmentation map in order to be able to discern pixels belonging to an object class (\textit{things}) from pixels belonging to non-object classes (\textit{stuff}).
In this paper, we built upon the proposed representation of \cite{Bai2017}, but overcome the requirement of an a-priorly known semantic segmentation by proposing a three-dimensional, instead of a two-dimensional, orientation map as intermediate instance representation, providing the flexibility to also represent non-object pixels by associating unit vectors pointing into the third dimension. 

\subsection{Semi-supervised segmentation}
Research on semi-supervised segmentation focusses on the question of how unlabelled data, which is typically easy to acquire in large amounts, can be used together with small amounts of labelled data to derive additional training signals in order to improve the segmentation performance.

One strategy for making use of unlabelled data is based on entropy minimisation \cite{Kalluri2019, Wittich2020}, where additional training signals are obtained by maximising the network's pixel-wise confidence scores of the most probable class using unlabelled data. 
However, this approach introduces biases for unbalanced class distributions, in which case, the model tends to increase the probability of the most frequent, and not necessarily of the correct classes.

In a semi-supervised segmentation setting using adversarial networks, the segmentation network is extended by a discriminator network that is added on top of the segmentation and which is trained to discriminate between the class labels being generated by the segmentation network and those representing the ground truth labels.
By minimising the adversarial loss, the segmentation network is enforced to generate predictions that are closer to the ground truth and, thus, they can be applied as additional training signals in order to improve the segmentation performance. 
In this context, the discrimination can be performed in an image-wise \cite{Luc2016} or pixel-wise \cite{Souly2017,Hung2018} manner. 
Since the adversarial loss can be computed without the need for reference labels once the discriminator is trained, the principles of adversarial segmentation learning are adapted for the semi-supervised setting to leverage the availability of unlabelled data \cite{Souly2017,Hung2018}.
Similar to the pixel-wise adversarial learning procedure of \cite{Souly2017,Hung2018}, the autors of \cite{Mendel2020} propose a correction network which is also added on top of the segmentation network and which learns on labelled data to distinguish between correct and incorrect class predictions. In the semi-supervised setting, the correction network is then used to produce additional supervision from unlabelled data based on the predictive certainty of the network.
However, learning the discriminator and the correction network, respectively, adds additional demands for labelled data and, therefore, may not reduce the need for such data in a way other strategies do. 

Closest to our approach is the line of research on semi-supervised segmentation based on the consensus principle. 
In this context, the authors of \cite{Ouali2020} train multiple auxiliary decoders on unlabelled data by enforcing consistency between the class predictions of the main and the auxiliary decoders.
Similarly, in \cite{Peng2020} two segmentation networks are trained via supervision on two disjunct datasets and additionally, by applying a co-learning scheme in which consistent predictions of both networks on unlabelled data are enforced. 
In \cite{Coenen2021a}, consistency training is additionally enriched by an auto-encoder branch, following the approach of auto-encoder regularisation \cite{Myronenko2019, Sedai2017} for semi-supervised learning. 
Another approach based on consensus training is presented in \cite{Xiaomeng2018, Zhang2020}, where unlabelled data is used in order to train a segmentation network by encouraging consistent predictions for the same input under different geometric transformations. 
While these approaches tackle the task of semantic segmentation, we extend the idea of consensus regularisation to a panoptic segmentation task by proposing the \textit{ConsInstancy} loss, which enforces consistency between semantic instance representations and semantic segmentation maps. 
In \cite{Hao2021}, a contour prior is introduced for instance-wise segmentation by assuming that the instance-segmentation boundaries should align with strong image gradients. 
However, expecting large image gradients at instance boundaries is a rather strong hypothesis which does not necessarily hold true for all scenes, particularly in the case of our fresh concrete data set (cf. Sec.~\ref{sec:datasets}).
Moreover, important to note is that in contrast to \cite{Hao2021} and other approaches for weakly supervised instance segmentation, as e.g.\;in \cite{Qizhu2018, Hsu2019} where weak annotations in the form of bounding boxes are needed for a weakly-supervised training, no additional annotations are required in our work.

\section{Methodology}
\subsection{Overview}
On an abstract level, encoder-decoder networks for \textbf{semantic segmentation} learn a function $f: X \rightarrow Y$ which maps the input images $X$ to pixelwise class predictions $Y$, such that $Y=\mathcal{D}^\text{seg}(\mathcal{E}(X))$.
In this setting, an encoder $\mathcal{E}(X)$ computes a latent feature embedding $\mathbf{z}$ from the input data and a segmentation decoder $\mathcal{D}^\text{seg}(\mathbf{z})$ is used to produce the label maps $Y$ from $\mathbf{z}$.  
Typically, $Y$ contains $i=1...N_C$ channels, one for each semantic class $C_i \in \mathbf{C}$, whose entries contain the pixelwise class-scores for the respective class $C_i$. 
In a \textbf{panoptic segmentation} setting, the semantic label set $\mathbf{C}$ distinguishes between subsets $\mathbf{C}^\text{St}$ and $\mathbf{C}^\text{Th}$, corresponding to \textit{stuff} and \textit{thing} classes, respectively, such that $\mathbf{C}=\mathbf{C}^\text{St}\cup \mathbf{C}^\text{Th}$.
In this definition, \textit{things} comprise classes of countable objects (here: the concrete \textit{aggregates}) whereas \textit{stuff} comprises classes of amorphous appearance and of similar texture or material (here: the cement \textit{suspension}).  
The task of panoptic segmentation extends the objective of semantic segmentation by mapping each pixel $p$ belonging to the subset $\mathbf{C}^\text{Th}$ not only to its semantic class $C_i^\text{Th}$ but in addition, also to an unique instance id $o \in \mathcal{O}$, enabling the differentiation of individual instances, with $\mathcal{O}$ denoting the set of all object instances.

Given a data set $X=\left\{X_l, X_u\right\}$, this paper presents a novel strategy to leverage unlabelled data $X_u$ along with labelled data $X_l$ for the training of a segmentation network in order to improve its performance compared to only using the labelled data.   
Specifically, we propose an \textit{instance decoder} $\mathcal{D}^\text{inst}$ which is trained to predict individual object class instances and which is added to the segmentation architecture while sharing the encoder $\mathcal{E}$ with the segmentation decoder $\mathcal{D}^\text{seg}$ (cf.\;Fig.~\ref{fig:overview}).
In this paper, we show that the proposed instance decoder serves multiple purposes.
On the one hand, by formulating the simultaneous prediction of semantic segmentation maps and instance representations as a multi-task framework, hence exploiting the complementary information of both disentangled but correlated tasks, the discriminative capability of the intermediate feature representations is improved and therefore leads to enhanced segmentation results. 
On the other hand, we demonstrate how to benefit from largely available unlabelled data, incorporating it into the training procedure by enforcing consistency between the predicted instance representations and the segmentation maps in order to produce additional self-supervised training signals and, thus, to significantly improve the performance of the network.
Lastly, we make use of the inferred instance representations to separate clustered objects, enabling a panoptic segmentation by generating instance-level semantic segmentation results.

\subsection{Semantic segmentation} \label{sec:segmentation}
Based on the latent feature embedding $\mathbf{z}$ produced by the encoder network $\mathcal{E}(X)$, the first output branch of our framework consists of a \textit{segmentation decoder} $\mathcal{D}^\text{seg}$ (cf. Fig.~\ref{fig:overview}) which predicts a pixelwise semantic segmentation $Y$. 

The encoder and \textit{segmentation decoder} architecture used in this work correspond to the \textit{\textbf{R}esidual depthwise \textbf{S}eparable convolutional \textbf{Net}work} (R-S-Net) proposed in \cite{Coenen2021a}.
This network consists of four encoder and decoder blocks, respectively. 
Each encoder-block consists of a residual convolution module, in which two intermediate representations are computed. 
The first representation is produced by a convolutional layer using a stride of 2, and the second one is computed by a sequence of a convolutional layer followed by a depthwise separable convolution layer and max-pooling. As output of the encoder-block, the elementwise sum of both intermediate representations is returned. 
Similar to that, the decoder-block processes the input in a two-stream path and returns the element-wise sum of the output of both streams. In the first
stream, the input is upsampled by a factor of 2, followed by a convolutional layer. The second stream consists of a sequence of one convolutional layer followed by a depthwise separable convolution and an upsampling layer. For more details, we refer the reader to \cite{Coenen2021a}. 

For the supervised training of the \textit{segmentation decoder}, labelled data $X_l$ and  associated reference labels are used to compute a pixelwise categorical cross-entropy loss $\mathcal{L}_\text{CE}$.

\subsection{Learning instance representations}
\begin{figure}[ht] 
\centering
\subfloat[Example image] {\includegraphics[width=0.32\columnwidth]{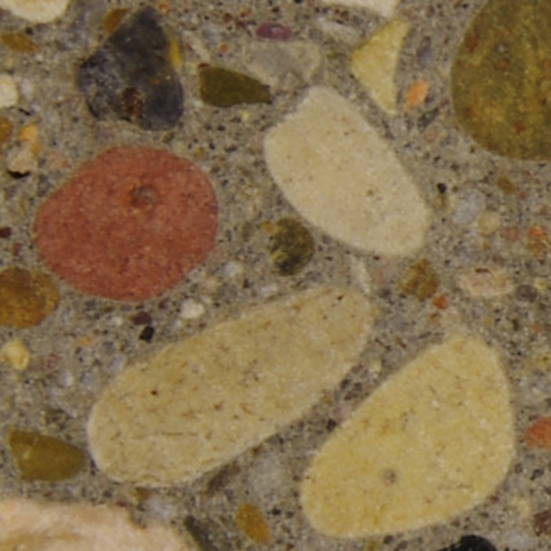}\label{fig:egImage}} 
\hspace{0.01cm}
\subfloat[Binary segmentation mask $Y$] {\includegraphics[width=0.32\columnwidth]{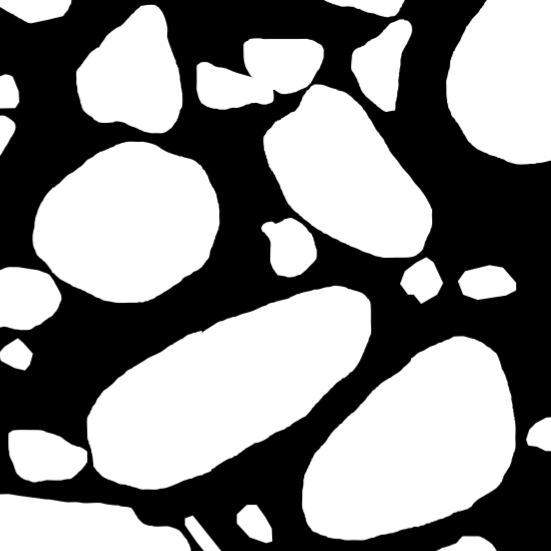}\label{fig:egSegmentation}} 
\hspace{0.01cm}
\subfloat[Instance boundaries] {\includegraphics[width=0.32\columnwidth]{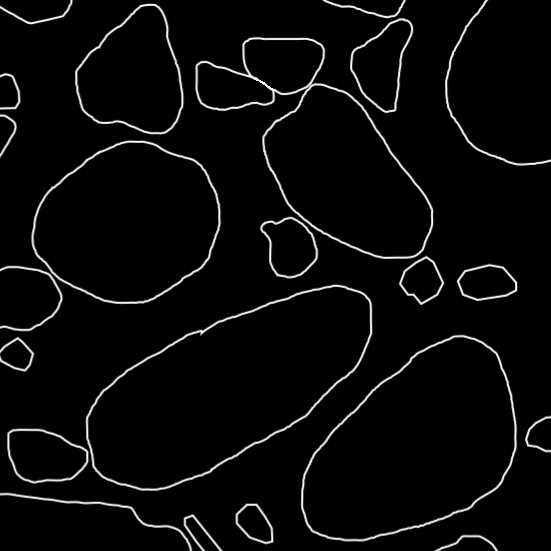}\label{fig:egBoundary}} 

\subfloat[Instance orientation map $\mathcal{I}^\theta$.] {\includegraphics[width=0.32\columnwidth]{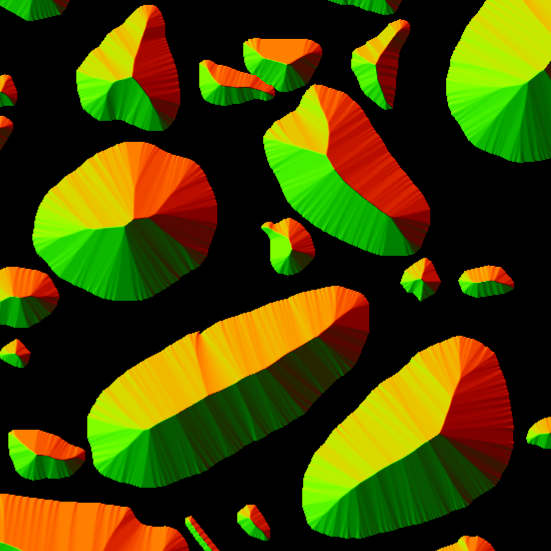}\label{fig:egOri}} 
\hspace{0.01cm}
\subfloat[Instance-aware dist. transform map $\mathcal{I}^{\delta}_{+}$.] {\includegraphics[width=0.32\columnwidth]{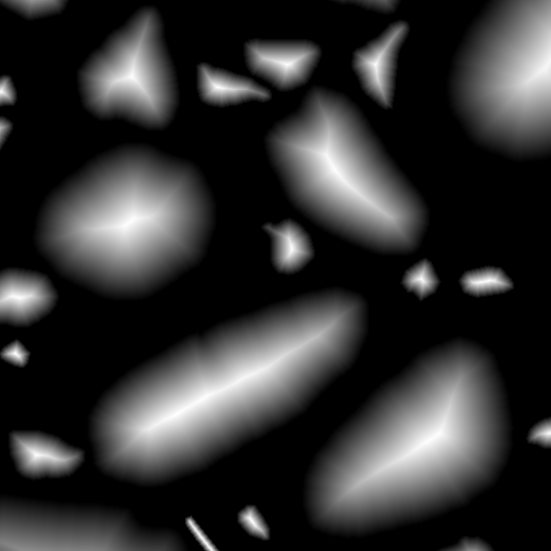}\label{fig:egDist}} 
\hspace{0.01cm}
\subfloat[Complem. distance transform map $\mathcal{I}^{\delta}_{-}$.] {\includegraphics[width=0.32\columnwidth]{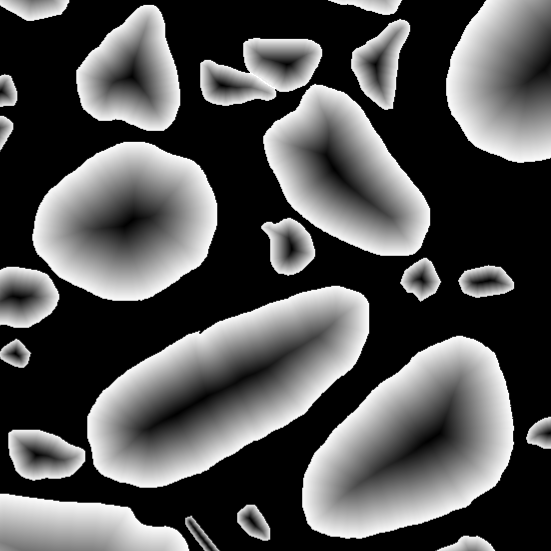}\label{fig:egDistInv}} 

\caption{Example images with its annotated segmentation and instance boundary masks (top row) together with their proposed instance representations (bottom row).}
\label{fig:egInstances} 
\end{figure}

In this section, we describe the instance representations that are proposed in this paper to represent instances of the \textit{thing} classes.
Examples of our representations are shown in Fig.~\ref{fig:egInstances}.
In particular, we train the \textit{instance decoder} $\mathcal{D}^\text{inst}$ to predict the two instance-aware distance transform maps $\mathcal{I}^{\delta}_{+}$ and $\mathcal{I}^{\delta}_{-}$ for each class in $\mathbf{C}^\text{Th}$ as final output (cf. Sec.~\ref{sec:distTrafo}).
Furthermore, we propose what we denote as \textit{instance orientation map} $\mathcal{I}^\theta$ as class-agnostic intermediate representation (cf. Sec.~\ref{sec:oriMap}). 
In our framework, this intermediate representation acts as additional guidance for predicting the final \textit{distance transform maps} (cf. Sec.~\ref{sec:instArchitecture}). 
%
%

\subsubsection{Distance transform maps} \label{sec:distTrafo}
The head of $\mathcal{D}^\text{inst}$ is designed to produce instance-aware distance transform maps. 
More specifically, as shown in Fig.~\ref{fig:overview}, we propose the prediction of two outputs.

On the one hand, the network predicts an instance-aware signed distance transform map $\mathcal{I}^{\delta}_{+}$ (cf.\ Fig.~\ref{fig:egDist}). 
Identical to the normal signed distance transform (SDT) \cite{Felzenswalb2012}, $\mathcal{I}^{\delta}_{+}$ represents the transformation of a binary segmentation mask $Y$ into an equivalent continuous representation.
However, while the original SDT assigns to each pixel of the foreground class its Euclidean distance to the closest point belonging to the background class, we define a slightly different representation by proposing an instance-aware SDT. 
In this representation, each foreground pixel gets assigned the Euclidean distance to its closest instance boundary point, i.e.\ the closest distance to either a background pixel or a pixel associated to another instance object. 
Pixels belonging to the background class are set to 0 in $\mathcal{I}^{\text{dist}}_{+}$.
In this way, we obtain an implicit representation of individual instances including the instance boundaries, in addition to the binary information, whether a pixel belongs to the semantic background or foreground class.
In comparison to a regular binary segmentation mask, in this representation, each pixel implicitly contains additional information about the spatial extent of its associated instance.   
In this way, the proposed representation enables the differentiation of individual instances even if they share a common boundary, which is not the case in the binary segmentation setting. 
As a side effect, we argue that this representation also helps the network to learn improved latent feature embeddings by being trained not only to predict a semantic class but also to discern individual instances at the same time.
Important to note is, that we perform an instance-wise normalisation of $\mathcal{I}^{\text{dist}}_{+}$, in which each entry is divided by the maximum Euclidean distance inherent to its associated instance. Formally, the entries $\delta_{p,+}$ of each foreground pixel $p$ in $\mathcal{I}^{\text{dist}}_{+}$ result in
\begin{equation}
\forall p\in o,o \in \mathcal{O}, \quad \delta_{p,+} = \frac{\min_{\overline{p} \notin o}(\lVert p-\overline{p}\rVert )}{\max_{\overline{p}\notin o}(\lVert p-\overline{p}\rVert)},
\end{equation}
where $o$ denotes an individual object instance within the set $\mathcal{O}$ of all instances and $\lVert p-\overline{p} \rVert$ returns the Euclidean distance between the instance pixel $p$ and the non-instance pixel $\overline{p}$.
By doing so, all values in $\mathcal{I}^{\delta}_{+}$ are in the range of $[0,1]$, which reduces the difficulty of using non-linear activation functions in order to model the scalar targets. 
Besides, the normalisation circumvents the effect that larger instances would contribute with a larger weight to the loss computation due to the appearance of larger Euclidean distances comprised by those objects compared to smaller instances.

As second output of the \textit{instance decoder} $\mathcal{D}^\text{inst}$, we introduce a complementary distance transform map $\mathcal{I}^{\delta}_{-}$.
This map is designed similar to $\mathcal{I}^{\delta}_{+}$, except that the values for each foreground pixel $p$ in $\mathcal{I}^{\delta}_{-}$ are set to $\delta_{p,-} = 1 - \delta_{p,+}$.
While the distance transform map $\mathcal{I}^{\delta}_{+}$ emphasises the skeleton of the instances and, therefore, has a rather poor representation of the instance boundaries due to the small contrast between background and foreground pixels at the low distance levels of the objects (cf.\ Fig.~\ref{fig:egDist}), the proposed complementary distance transform map $\mathcal{I}^{\delta}_{-}$ introduces strong training signals for background-foreground confusions in instance boundary regions (cf.\ Fig.~\ref{fig:egDistInv}).
This representation, therefore, encourages the network to learn an accurate estimation of the instance contours. 

Per definition, the relationship of the distance transform maps $\mathcal{I}^{\delta}_{+}$ and $\mathcal{I}^{\delta}_{-}$ and the binary label map $Y$ results in
\begin{equation}
Y = \mathcal{I}^{\delta}_{+} + \mathcal{I}^{\delta}_{-}. \label{eq:relationship}
\end{equation}
In the semi-supervised setting proposed in this paper, this relationship is exploited in order to enforce consistency between the instance predictions and the segmentation results of our framework (cf. Sec.~\ref{sec:semisupervised}).

\subsubsection{Instance orientation map} \label{sec:oriMap}
\begin{figure}[ht]
\begin{center}
		\includegraphics[width=1.0\columnwidth]{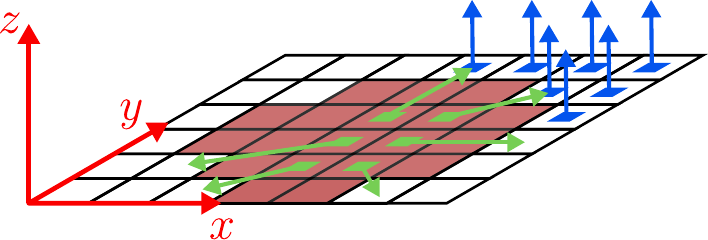}
	\caption{Schematic visualisation of the instance orientation map $\mathcal{I}^\theta$. Pixels belonging to an instance are coloured in red. The green and blue arrows indicate the 3D unit vectors $\theta_p$.}
\label{fig:OriMap}
\end{center}
\end{figure}

Inspired by \cite{Bai2017}, we design $\mathcal{D}^\text{inst}$ to predict an \textit{instance orientation map} $\mathcal{I}^\theta$ as an intermediate instance representation. 
In this representation, each pixel $p$ of the input image is parametrized by a three-dimensional unit vector $\theta_p$ as depicted in Fig.~\ref{fig:OriMap}. 
For pixels $p$ not belonging to an object instance we define $\theta_p$ as a unit vector perpendicular to the image plane. 
For pixels $p$ belonging to one of the object instances $\mathcal{O}$, $\theta_p$ corresponds to the unit vector in the image plane pointing towards the closest pixel not belonging to the respective instance.
Thus, according to this definition, $\theta_p$ of each instance pixel corresponds to the normalized gradient of the instance distance transform map $\mathcal{I}^\delta_+$ at the corresponding pixel position so that 
\begin{equation}
\theta_p =
\begin{cases}
\left[\frac{\nabla \mathcal{I}^\delta_+}{\lVert \nabla \mathcal{I}^\delta_+ \rVert},\ 0\right]^{T} & \text{for } p \in \mathcal{O} \\
\left[0,0,1\right]^T & \text{for } p \notin \mathcal{O}.
\end{cases}
\end{equation}
In this equation, $\nabla\mathcal{I}^\delta_+ = [\nabla x, \nabla y]$ denotes the two-dimensional gradient vector containing the gradient components in the x and y direction in image space, respectively, and $\lVert \nabla \mathcal{I}^\delta_+ \rVert$ denotes the norm of the gradient. 

Note that our three-dimensional representation of the \textit{instance orientation map} is different from the one applied in \cite{Bai2017}, where the proposed \textit{direction network} produces a map of only two-dimensional in-plane unit vectors. 
In that case, meaningful directional vectors can only be defined for instance-pixels and therefore, a semantic segmentation of the input is required beforehand in order to differentiate between instance and non-instance image regions. 
In this paper, the information whether a pixel belongs to an object instance (in-plane unit vector) or to a non-object class (out-of-plane unit vector) is implicitly encoded in the three-dimensional orientation field.
As a consequence, in our framework, a prior segmentation of the input as it is done by \cite{Bai2017} is not required in order to predict the instance orientation maps. 
In contrast, we argue that by implicitly encoding the semantic information of the pixel associations to either the object or non-object class in the orientation map, we force the network to learn to distinguish between these cases, enforcing the extraction of richer and more discriminative features. 
Furthermore, we leverage the property of the instance orientation map to possess perpendicular vectors at the boundary between \textit{thing} instances and \textit{stuff} regions as well as opposed vectors (i.e.\ maximum angular differences) at boundary pixels between two neighbouring instance objects.
We argue, that, in this way, we enforce the network to learn very accurate instance boundary localisations at pixel level. 
In accordance to \cite{Bai2017}, we believe that learning the proposed orientation map $\mathcal{I}^\theta$, as an intermediate representation for the instance landscape of the input image, aids the instance-decoder $\mathcal{D}^\text{inst}$ in producing the distance transform maps as the final output.

\subsubsection{Instance decoder architecture} \label{sec:instArchitecture}
\begin{figure}[ht]
\begin{center}
		\includegraphics[width=0.7\columnwidth]{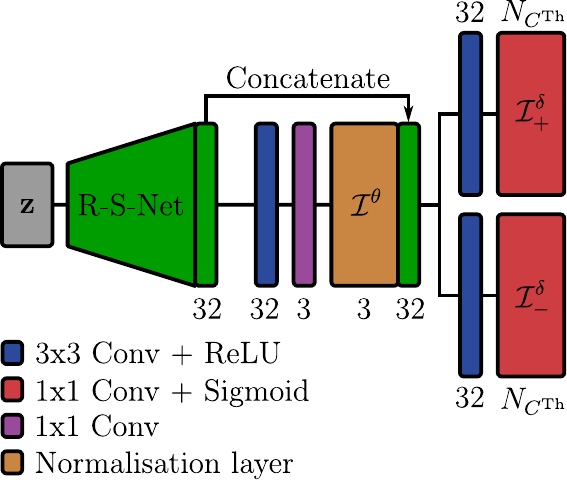}
	\caption{Architecture of the \textit{instance decoder} $\mathcal{D}^\text{inst}$.}
\label{fig:InstDecoder}
\end{center}
\end{figure}

An overview on the architecture which is employed for the instance decoder $\mathcal{D}^\text{inst}$ is depicted in Fig.~\ref{fig:InstDecoder}. 
The input to the decoder is the shared feature embedding $\mathbf{z}$ produced by the encoder $\mathcal{E}$ for the input image $X$. 
As the \textit{segmentation decoder}, the \textit{instance decoder} backbone applies the decoder of the R-S-Net \cite{Coenen2021a}.
The backbone produces a feature map of depth 32 and of the same resolution as the input image.
A 3x3 convolutional layer with ReLU activation, followed by a 1x1 convolutional layer with no activation, is used to predict a three-channel feature map, allowing values in the range $[-\infty, \infty]$.
Subsequently, in order to produce the orientation map $\mathcal{I}^\theta$, a normalisation layer is applied in order to restrict the pixelwise sum of each channel’s squared output to  1, thus, forming the three-dimensional unit vectors $\theta_p$ as shown in Fig.~\ref{fig:OriMap}. It has to be noted, that the orientation map is class agnostic, i.e.\ we design the network to produce one orientation map representing all instances, regardless of their associated classes. 
The prediction of the final distance transform maps takes place considering the predicted orientation maps by leveraging the orientation maps concatenated with the ouput feature map of the backbone decoder as input. A 1x1 convolutional layer with sigmoid activation is applied to produce the transform maps $\mathcal{I}^{\delta}_+$ and $\mathcal{I}^{\delta}_-$ comprising a total of $N_{C^\text{Th}}$ channels, one for each class in $\mathbf{C}^\text{Th}$.\\

\textbf{Training:} In order to train the \textit{instance decoder} in a supervised manner, reference data for the orientation map and the distance transform maps are required.
Given the labelled data $X_l$, i.e.\ the reference semantic instance segmentation masks, these reference maps can directly be computed from the instance masks, though, and do not lead to additional labelling requirements. 
We define the loss for the orientation map $\mathcal{I}^\theta$ in the angular domain by applying the cosine similarity loss $\mathcal{L}_\text{COS}$ and make use of the mean squared error (MSE) as loss $\mathcal{L}_\text{MSE,+}$ and $\mathcal{L}_\text{MSE,-}$ for the output of the distance transform maps $\mathcal{I}^{\delta}_+$ and $\mathcal{I}^{\delta}_-$, respectively.

\subsection{Panoptic segmentation} \label{sec:panoptic}
While a semantic segmentation of the input is delivered directly by the network as output of the \textit{segmentation decoder} (cf.\ Sec.~\ref{sec:segmentation}), deriving the panoptic segmentation using the proposed framework requires post-processing. 
In a first step, we make use of the predicted \textit{complementary distance transform map} $\mathcal{I}^\delta_-$ in order to extract the outlines for the instances of each semantic class in $\mathbf{C}^\text{Th}$ by thresholding $\mathcal{I}^\delta_-$ using a threshold of $0.9$.
Note that this is in contrast to \cite{Bai2017}, where the low distance areas of a regular distance transform map are used as energy cut, which, however, does not define the instance boundaries as well as compared to the \textit{complementary distance transform map} proposed in this paper.
In a second step, we subtract the extracted instance outline map from the binary semantic segmentation map of the corresponding class in $Y$, and subsequently, associate each remaining connected component with an individual instance ID. 
Finally, instance boundary pixels are allocated to the id of their neighbouring instance, resulting in a  panoptic segmentation of the input image.
 
\subsection{Semi-supervised training} \label{sec:semisupervised}
In this paper, we propose a strategy to incorporate unlabelled data $X_u$, in addition to the limited amount of labelled data $X_l$, to the training procedure of our segmentation network in order to improve its performance. 
To this end, we define the overall training objective as to minimise the overall training loss $\mathcal{L}$ with
\begin{equation}
\mathcal{L} = \underbrace{\mathcal{L}_\text{CE} + \mathcal{L}_\text{COS} + \mathcal{L}_\text{MSE,+} + \mathcal{L}_\text{MSE,-}}_{\text{supervised}} \ + \underbrace{\mathcal{L}_\text{CONS}}_{\text{unsupervised}}, \label{eq:lossfunction}
\end{equation}
being composed of supervised loss functions which require the availability of reference data and an unsupervised loss function which does not require any reference data to be available. 
The supervised loss functions produce training signals at the outputs of the network for the semantic segmentation mask, the orientation map, and the distance transform maps, respectively. These signals are based on the discrepancy between the predictions and the provided reference data and are computed according to the loss functions described in Sec.~\ref{sec:segmentation} and Sec.~\ref{sec:instArchitecture}.
We would like to point out that in order to compute the supervised loss functions, no additional annotations other than the instance-level annotations are required. 
Instead, the reference data required for the individual loss terms, i.e.\;the different instance representations, can directly be derived from the instance-level annotations. 
As a consequence, our proposed method does not add further annotation efforts but instead, makes use of different representations of existing annotations at instance level to enrich the training procedure by formulating the supervised part of the total loss in Eq.~\ref{eq:lossfunction} as a multi-task learning problem.

In order to compute the unsupervised loss from unlabelled data, we propose the \textbf{ConsInstancy} loss which we define as
\begin{equation}
\mathcal{L}_\text{CONS} = \sum_{i=1}^{N_{C^\text{Th}}} \mathcal{L}_\text{MSE}(Y_i, \mathcal{I}^\delta_{+,i}+\mathcal{I}^\delta_{-,i}).
\end{equation}
In this definition, we make use of the relationship described by Eq.~\ref{eq:relationship}, namely that the sum of the two predicted distance transform maps 
$\mathcal{I}^\delta_{+,i}$ and $\mathcal{I}^\delta_{-,i}$ for the \textit{thing} classes $C_i \in \mathbf{C}^\text{Th}$ must result in the binary label representation $Y_i$ predicted for that class.
This relationship allows us to introduce the MSE between the predicted label maps and the sum of the predicted distance transform maps as additional unsupervised training signals, derived from entirely unlabelled data (cf.\ Fig.\ref{fig:overview}). 
By minimising this loss based on the discrepancy between the individual outputs, we enforce consistency between the predictions of the \textit{segmentation decoder} and the \textit{instance decoder}, enabling the exploitation of the consensus principle \cite{Chao2016}.
This principle is founded on the rationale that enforcing an agreement between different outputs of the same network restricts the parameter search space to cross-consistent solutions and, therefore, acts as additional regularisation on the shared encoder, thus, enhancing its feature representation and improving its generalisation ability.
A high-level overview on the proposed framework is shown in Fig.~\ref{fig:overview} and an overview on the training procedure of the proposed semi-supervised segmentation approach is given in Alg.~\ref{alg:consinstancy}.

\begin{algorithm}
\caption{Semi-supervised panoptic segmentation}\label{alg:consinstancy}
\begin{algorithmic}[1]

\State \textbf{Input:} Data set $X=\left\{X_l, X_u\right\}$ and labels $Y_l$
\Procedure{ConsInstancy Training}{}
\State Setup network architecture
\State Initialise network weights $\mathbf{w}$
\For{i in number of epochs}
\State Predict $\hat{Y},\ \mathcal{I^\theta},\ \mathcal{I}^\delta_+,\ \mathcal{I}^\delta_-$ for $X_l$ 
\State Predict $\hat{Y},\ \mathcal{I^\theta},\ \mathcal{I}^\delta_+,\ \mathcal{I}^\delta_-$ for $X_u$
\State Compute supervised losses 
\State Compute unsupervised loss ($\mathcal{L}_\text{CONS}$)
\State Compute total loss $\mathcal{L}(\mathbf{w})$
\State Update weights $\mathbf{w} = \mathbf{w}-\eta\nabla\mathcal{L}(\mathbf{w})$
\EndFor

\EndProcedure 

\end{algorithmic}
\end{algorithm}

\section{Experimental evaluation}
In this section, we evaluate our approach on two semantic instance-level data sets of concrete aggregates.
We analyse the performance on both, semantic segmentation and panoptic segmentation tasks. In this context, we also perform ablation studies of our proposed method in order to examine the impact of the different constituents of our model. 

\subsection{Test data} \label{sec:datasets}
\begin{figure}[ht]
\begin{center}
		\includegraphics[width=0.95\columnwidth]{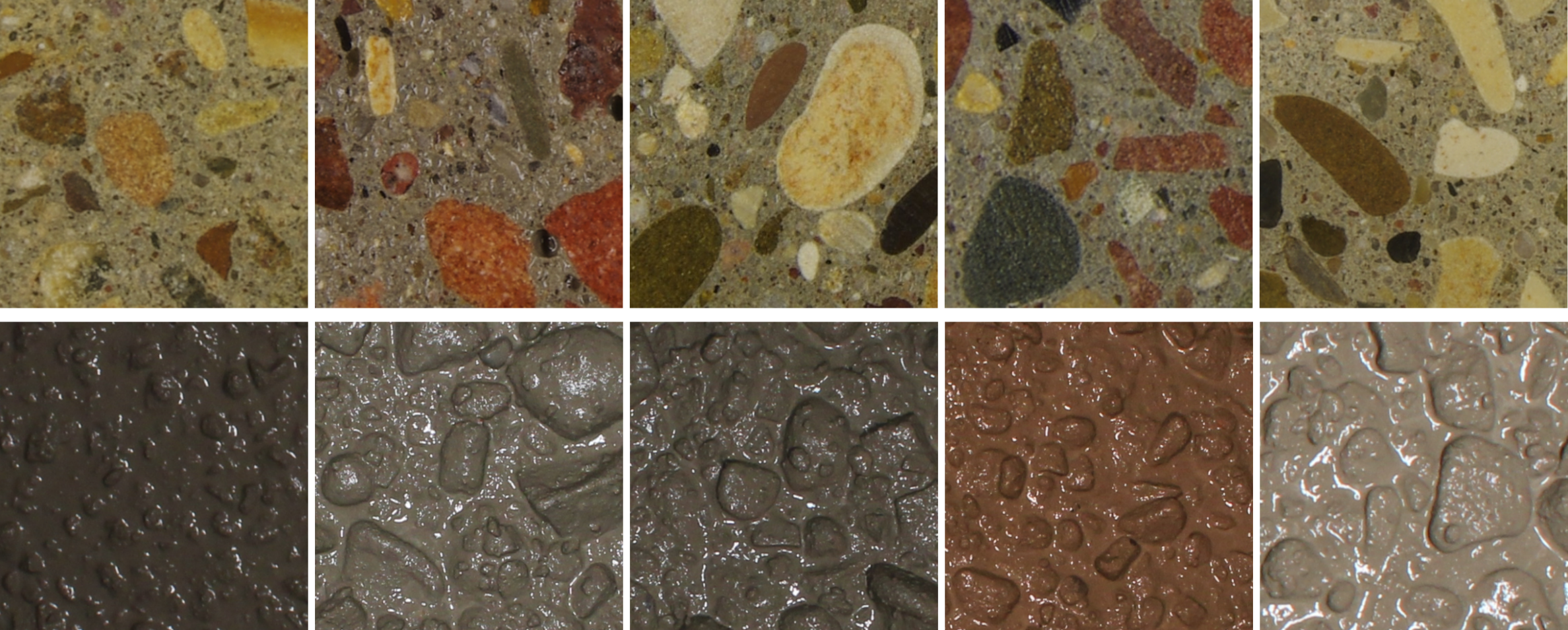}
	\caption{Exemplary images of the two data sets used for evaluation in this work. Top row: images of the \textit{sedimentation} data set \cite{Coenen2021a}, depicting aggregate particles in hardened concrete. Bottom row: images of our proposed \textit{fresh concrete} data set, showing aggregate particles in fresh concrete.}
\label{fig:Dataset2}
\end{center}
\end{figure}

We experimentally evaluate our proposed method on two different data sets of concrete aggregates.  
Both data sets used in this work distinguish the classes \textit{suspension (stuff)} and \textit{aggregate (thing)}.
The first data set is the concrete \textit{sedimentation} data set proposed in \cite{Coenen2021a}. 
It consists of 612 labelled and 827 unlabelled image tiles of hardened and lengthwise cut concrete cylinders with a resoultion of 448x448$\,\text{px}^2$. 
Exemplary tiles of the sedimentation data are shown in the top row of Fig.~\ref{fig:Dataset2}.

\begin{figure}[ht]
\begin{center}
		\includegraphics[width=0.95\columnwidth]{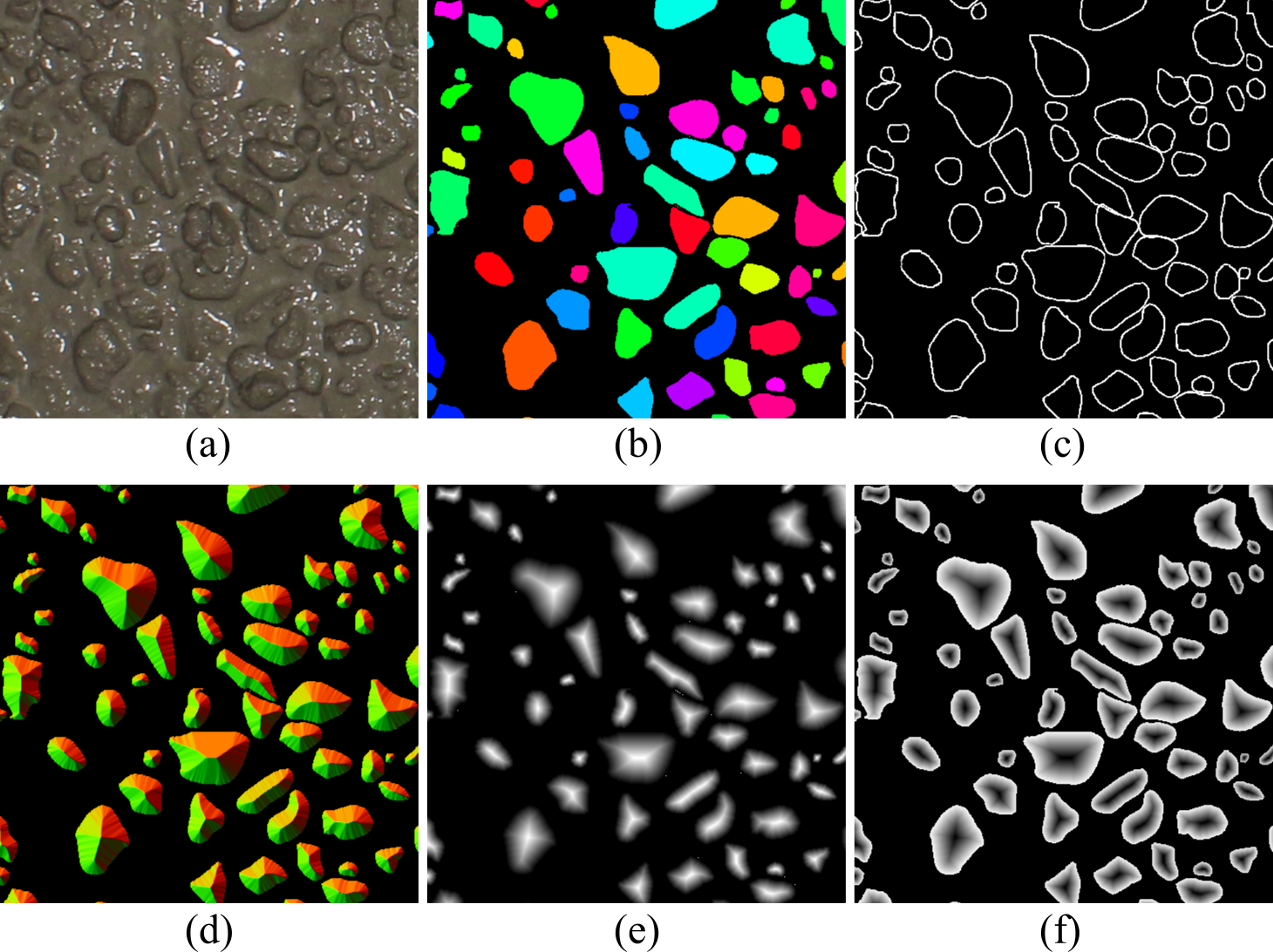}
	\caption{Exemplary image of the \textit{fresh concrete} data set \textbf{(a)} and its corresponding reference maps: The instance-level segmentation mask \textbf{(b)}, the instance boundaries \textbf{(c)},  the orientation map \textbf{(d)} as well as the distance transform map \textbf{(e)} and  its complementary representation \textbf{(f)} are shown.}
\label{fig:Dataset}
\end{center}
\end{figure}

In contrast, the second data set contains images of \textit{fresh concrete} acquired and annotated by ourselves during the standard approach for quality control of fresh concrete at construction sites, the so called slump test \cite{Gambhir2004}. 
It contains 1096 images of size 480x480$\,\text{px}^2$, manually labelled at instance-level. Furthermore, we made use of an additional set of 2000 unlabelled images for the semi-supervised training in our experiments. 
Exemplary tiles of our fresh concrete data are shown in the bottom row of Fig.~\ref{fig:Dataset2} and an example image, together with its corresponding reference maps, is depicted in Fig.~\ref{fig:Dataset}.
From Fig.~\ref{fig:Dataset2}, the diversity of the appearance of both, aggregate and suspension in both data sets can be noted.
In comparison to the \textit{sedimentation} data obtained on hardened concrete, the \textit{fresh concrete} data is more challenging as the particles are embedded in the viscous binder suspension, rendering parts of the instance boundaries indistinct and ambiguous.
 
\subsection{Test setup}
\textbf{Ablation studies:}
To assess the effect of the individual constituents of our framework, we perform tests using different network variants, considering different components in the evaluation. 
In the \textbf{Seg} setting, we train a semantic segmentation network by only using the \textit{segmentation decoder}, while disregarding the \textit{instance decoder} during training. Consequently, in this setting, we perform a purely supervised training without the incorporation of unlabelled data.
In the \textbf{Inst} variant, we perform training using both, the \textit{segmentation} and the \textit{instance decoder}, but again, we only perform a purely supervised training. 
Thus, with this setting, the effect of multi-task learning, which is achieved by not only training the network for semantic segmentation, but at the same time for predicting the instance representations, can be analysed. 
In the \textbf{ConsInst} variant, we make use of our complete approach proposed for the semi-supervised panoptic segmentation. 
Thus, in this setting, the unlabelled data is leveraged by computing the \textit{ConsInstancy} loss for a semi-supervised training. \\

\textbf{Training:} 
The networks used in the different variants of the proposed framework are trained from scratch.
The convolutional layers are initialised using the \textit{He} initialiser \cite{He2015}.
The networks are trained using the Adam optimizer \cite{Adam2015}, using the exponential decay rate for the $1^\text{st}$ moment estimates $\beta_1 = 0.9$ and for the $2^\text{nd}$ moment estimates $\beta_2 = 0.999$.  
We apply weight regularisation on the convolutional layers using L2 penalty with a regularisation factor of $10^{-5}$.
A mini-batch size of 8 is applied, meaning that in the semi-supervised setting, each mini-batch consists of four labelled and four unlabelled training images. 
We use an initial learning rate of $ 10^{-3}$ and decrease the rate by a factor of $10^{-1}$ after 25 epochs with no improvement in the training loss.
In all settings, we make use of the same, very limited amount of labelled data for training. 
In case of the \textit{sedimentation} data, we use only 17 labelled images and all unlabelled images for training, as suggested in \cite{Coenen2021a}. 
In case of our \textit{fresh concrete} data set, we considered 150 labelled images for training of the supervised components of the framework, and all unlabelled images for the semi-supervised part. \\

\textbf{Metrics:}
The evaluation is carried out based on all annotated images that are not used for training.
We report results for different evaluation metrics. 
In order to evaluate the performance of the approach related to the pixelwise semantic segmentation, we determine values for the class-wise recall (R), precision (P) and $F_1$ scores as well as the overall segmentation accuracy (OA). Additionally, since the OA can be biased towards more frequent classes, we report the mean $F_1$ score  of the segmentation ($MF_1^\text{seg}$), computed as average of the class-wise $F_1$ score of all classes. 
Furthermore, we analyse the performance of the approach towards instance- and panoptic segmentation and report results for the instance-wise $F_1^\text{inst}$ score to assess the performance on instance-level segmentation and results for \textit{panoptic quality} (PQ), the metric for panoptic segmentation defined in \cite{Kirillov2019} with
\begin{equation}
PQ = \frac{\sum_{(p,g)\in TP} \text{IoU}(p,g)}{\lvert TP\rvert+\frac{1}{2}\lvert FP\rvert+\frac{1}{2}\lvert FN \rvert}.
\end{equation}
In this context, $p$ and $g$ denote predicted and ground truth segments, respectively. Furthermore, $TP$ (true positive) denotes correctly matched instances, $FP$ (false positive) and $FN$ (false negative) represent unmatched predicted and ground truth instances, respectively. 
For both, $F_1^\text{inst}$ and PQ metrics, we require the segmentation masks of instances to have an intersection-over-union (IoU) of 50\% or more with a reference instance mask to be counted as true positive segmentation. \\

\textbf{Comparison to state-of-the-art:}
We compare our results with the results achieved by two current state-of-the-art approaches for semi-supervised segmentation of \cite{Coenen2021a} and \cite{Ouali2020}. 
To enable a fair comparison, we trained both approaches from scratch using the same labelled and unlabelled data that was used for the training of the proposed framework.
The results achieved for the task of semantic segmentation can directly be compared to the results obtained by our approach. 
In order to compute the evaluation metrics for the panoptic segmentation, we identify connected components in the semantic label space. From these components we derive an instance-wise segmentation, since both state-of-the-art methods only deliver a semantic but no panoptic segmentation of the input. 
Note, that the same is true for the defined \textit{Seg} variant of our framework, where only the segmentation branch is applied. 
We point out that a comparison of the panoptic segmentation metrics consequently is not one-hundred percent fair. 
However, in case of the \textit{sedimentation} data, the effect that adjacent particles have on an instance-wise evaluation is almost negligible since an overlap of multiple particles in the profiles of the concrete cylinders is physically impossible and the occurrence of directly adjacent particles is very rare.
Nevertheless, in case of the \textit{fresh concrete} data set, the effect of identifying instances from connected components is an issue and, therefore, impacts the metrics for the panoptic segmentation of the approaches, which is why the comparison in Tab.~\ref{tab:resultsInst} have to be taken with caution. Still, we decided to include the results in this paper, as they also show the contribution of our framework of extending approaches for semi-supervised segmentation by also predicting instance masks in addition to the semantic segmentation mask.

\subsection{Results}
In this section, we evaluate the results achieved by the proposed approach for semi-supervised panoptic segmentation on the two described data sets. 
In this context, we analyse the effects of the individual components of the approach, nameley the prediction of the proposed instance representations and the usage of the \textit{ConsInstancy} training.

\subsubsection{Semantic segmentation:}
Tab.~\ref{tab:resultsSegClasswise} (class-wise evaluation scores) and Tab.~\ref{tab:resultsSeg} (overall accuracy and $MF_1$ scores) contain the results for the metrics chosen to assess the quality w.r.t.\ to the performance on the semantic segmentation task.

\begin{table*}[ht]
	\centering
	\caption{Class-wise scores for recall (R), precision (P) and $F_1$ for the pixelwise semantic segmentation achieved on both data sets for the classes \textit{suspension} and \textit{aggregate}. The maximum achieved $F_1$ score is depicted in bold.}
		\begin{tabular}{|l| c c c| c c c || c c c | c c c  |} \hline
		 &\multicolumn{6}{c||}{\textit{Sedimentation}} & \multicolumn{6}{c|}{\textit{Fresh concrete}}\\
		Class-wise scores & \multicolumn{3}{c|}{Suspension} & \multicolumn{3}{c||}{Aggregate} & \multicolumn{3}{c|}{Suspension} & \multicolumn{3}{c|}{Aggregate} \\		
		in [\%]& R & P & $F_1$ & R & P & $F_1$ & R & P & $F_1$ & R & P & $F_1$\\ \hline
		Ours (\textbf{Seg}) & 87.2 & 89.6 & 88.4 & 81.1 & 77.3 & 79.2 & 95.8 & 95.5 & 95.6 & 68.7 & 70.3 & 69.5 \\
		Ours (\textbf{Inst}) & 89.3 & 90.9 & 90.1 & 83.5 & 80.7 & 82.1 & 96.7 & 95.1 & 95.9 & 65.9 & 74.1 & 69.7 \\
		Ours (\textbf{ConsInst}) & 95.7 & 89.7 & \textbf{92.6} & 79.6 & 90.9 &\textbf{84.9} & 97.0 & 95.6 & \textbf{96.3} & 69.6 & 76.8 & \textbf{73.0}\\ \hline
		Ouali et al. (2020)\cite{Ouali2020} & 98.3 & 86.1 & 91.8 & 70.4 & 95.7& 81.2 & 97.1 & 95.2 & 96.1 & 66.4 & 76.6 & 71.2 \\ 
		Coenen et al. (2021) \cite{Coenen2021a}	& 94.8 & 88.5 & 91.5 & 77.2 & 88.8 & 82.6 & 96.9 & 95.4 & 96.2 & 68.0 & 76.3 & 71.3 \\ \hline
		\end{tabular}	
\label{tab:resultsSegClasswise}
\end{table*}  

\begin{table}[ht]
	\centering
	\caption{Overall accuracies and $MF_1$ scores for the pixelwise semantic segmentation achieved on both data sets.}
		\begin{tabular}{|l| c c | c c |} \hline
		&\multicolumn{2}{c|}{\textit{Sedimentation}} & \multicolumn{2}{c|}{\textit{Fresh concrete}}\\
		in [\%]& OA & $MF_1^\text{seg}$ & OA & $MF_1^\text{seg}$ \\\hline		
		Ours (\textbf{Seg})  							& 85.1 & 83.8 & 92.4 & 82.6 \\
		Ours (\textbf{Inst}) 							& 87.2 & 86.1 & 92.8 & 82.8 \\
		Ours (\textbf{ConsInst})  				& \textbf{90.1} & \textbf{88.7} & \textbf{93.5} & \textbf{84.7}\\ \hline
		Ouali et al. (2020) \cite{Ouali2020} 		& 88.5 & 86.5 & 93.2 & 83.6 \\ 
		Coenen et al. (2021) \cite{Coenen2021a}	& 88.6 & 87.1 & 93.3 & 84.1 \\ \hline
		\end{tabular}	
\label{tab:resultsSeg}
\end{table}  

To enable a better evaluation of the performance metrics achieved by the different network variants, we conduct a sensitivity analysis in order to assess the performance variations of the variants resulting from different weight initialisations and training.
We do this on the example of the \textbf{Inst} and \textbf{ConsInst} models and the \textit{sedimentation} data, by training these models multiple (four) times from scratch, using a different weight initialisation each time, and by computing the average and standard deviations of the segmentation quality metrics achieved by the models (Tab.~\ref{tab:sensitivity}).  
\begin{table}[ht]
	\centering
	\caption{Sensitivity analysis. Standard deviations for different quality metrics obtained by training the models multiple times from scratch, using different weight initialisations, on the \textit{sedimentation} data set.}
		\begin{tabular}{|l| c c | c c |} \hline
		Std. dev.& OA & $MF_1^\text{seg}$ & $F_1$ & $F_1$ \\ 
		in [\%]& & & (suspension) & (aggregate) \\ \hline
		\textbf{Inst} & 0.48 & 0.45 & 0.45 & 0.48\\
		\textbf{ConsInst} & 0.47& 0.48 & 0.41 & 0.64\\ \hline
		\end{tabular}	
\label{tab:sensitivity}
\end{table}  
As can be seen from the table, the resulting standard deviations of the quality metrics are relatively small, namely almost exclusively less than half a percent (an exception is the $F_1$ score for the class \textit{aggregate}, with 0.64\%). 
In the following ablation analysis, these values deliver indications for the assessment whether differences between the performance of the analysed model variations are significant or not.

The \textbf{Seg} variant, i.e.\ the semantic segmentation network without the \textit{instance decoder} trained in a purely supervised manner, achieves an OA of 85.1 and 92.4\% and a $MF_1^\text{seg}$ score of 83.8 and 82.6\% on the two evaluated data sets, respectively (cf. Tab.~\ref{tab:resultsSeg}). 

The \textbf{Inst} variant adds the proposed \textit{instance decoder} during training and, thus, performs multi-task learning but still does not use any unlabelled data. 
As can be seen from the results in Tab.~\ref{tab:resultsSegClasswise}, only by applying multi-task learning using our proposed instance representations, the class-wise metrics including the $F_1$ scores increase for both classes on the \textit{sedimentation} data set by up to 2.9\%.
Consequently, also the OA and $MF_1^\text{seg}$ score increase (+2.1\% and +2.3\%, respectively).
Compared to the performance variations of the individual models which are shown in Tab.~\ref{tab:sensitivity} (less than 0.5\% of standard deviations), these improvements are distinctly larger than the $3\sigma$ interval and, therefore, can be evaluated as significant. 
On the \textit{fresh concrete} data set, improvements are also visible but less distinct.
It can be noted, that especially the precision of the class \textit{aggregate} (i.e.\,the \textit{thing} class) benefits from the consideration of the proposed instance decoder. 
Compared to the \textbf{Seg} variant, the precision for that class increases by 3.4\% on the \textit{sedimentatio} data and by 3.8\% on the \textit{fresh concrete} data. 
We argue, that enforcing the network to explicitly learn the discrimination of individual instances within certain classes leads to the extraction of more discriminant features, therefore enabling a more precise classification of the respective classes.

\begin{figure}[ht]
\begin{center}
		\includegraphics[width=0.99\columnwidth]{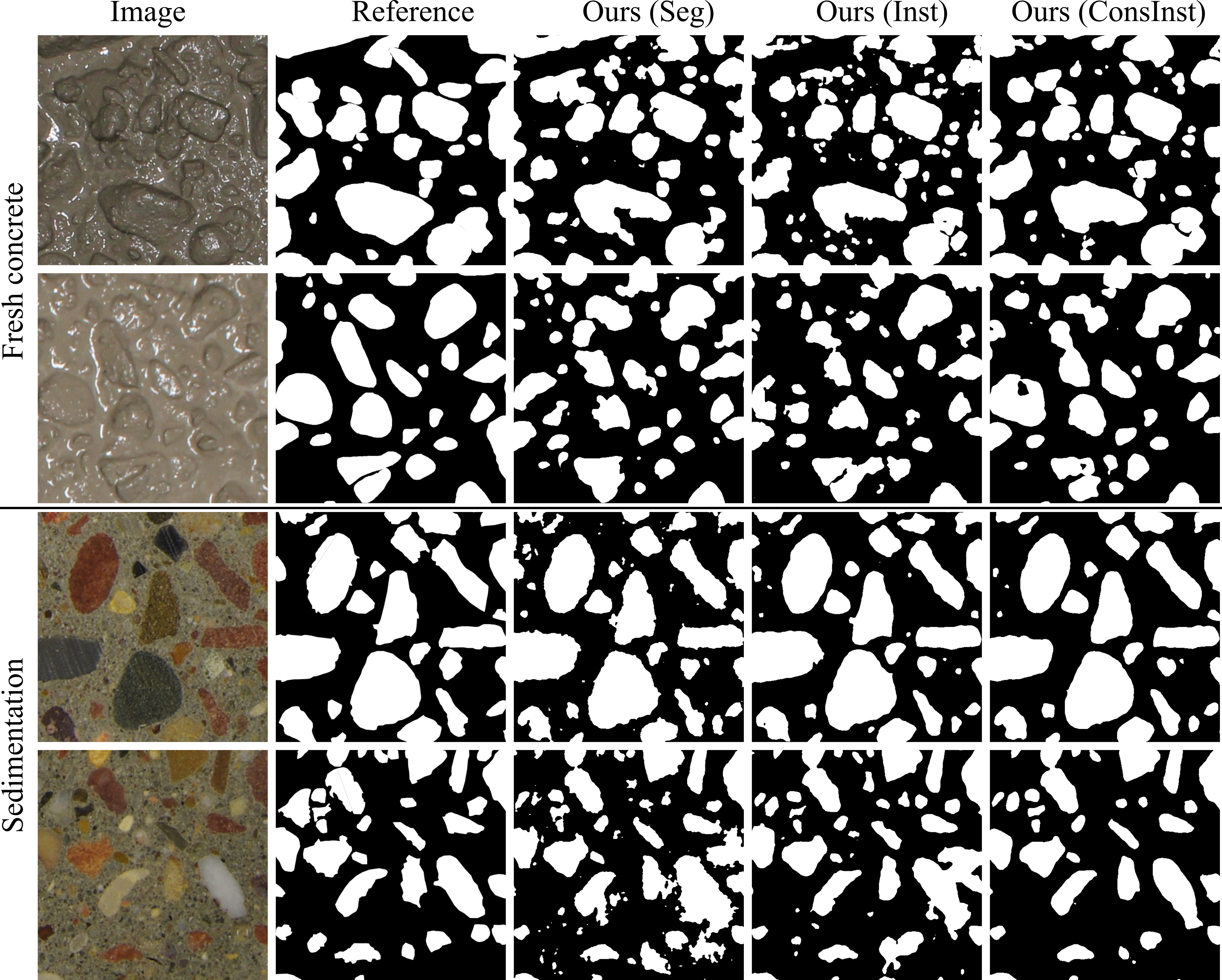}
	\caption{Qualitative results for the segmentation masks obtained by the variants used in the ablation studies of our proposed framework.}
\label{fig:QualRes}
\end{center}
\end{figure}

Making use of unlabelled data in the \textbf{ConsInst} setting additionally to the labelled data during training, by applying our proposed \textit{ConsInstancy} training, again, the performance of the semantic segmentation is enhanced by a margin for all metrics on both data sets. The $MF_1^\text{seg}$ scores for instance increase by a significant margin of 2.6\% and 1.9\%, respectively, demonstrating the potential of our framework.
As is visible from Tab.~\ref{tab:resultsSegClasswise}, our \textit{ConsInstancy} training, again, particularly favours the segmentation of the \textit{thing} related pixels of the class \textit{aggregate}, as the largest improvements for the $F_1$ score are achieved for that class.
For a visual comparison of the segmentation performance, qualitative results for the segmentation masks obtained by the evaluated variants of our framework are shown in Fig.~\ref{fig:QualRes}.
As is visible, compared to the \textbf{Seg} and the \textbf{Inst} variant, the \textbf{ConsInst} setting especially leads to distinctly smoother segmentations of the instance boundaries and to a significant reduction of spurious and erroneous  false positive elements on both data sets. 
This visual impression is quantitatively supported by the values of \textit{precision} achieved for the class \textit{aggregate} (cf. Tab.~\ref{tab:resultsSegClasswise}), which are distinctly enhanced by the \textbf{ConsInst} variant, indicating the significant reduction of the mentioned false positive segmentations of that class. 

Compared to the state-of-the-art methods of \cite{Ouali2020} and \cite{Coenen2021a}, whose results are also reported in Tab.~\ref{tab:resultsSeg}, our semi-supervised approach achieves superior results for both, the OA and the $MF_1^\text{seg}$ score, on both data sets. (Note that inconsistencies between the numbers reported here for \cite{Coenen2021a} on the \textit{sedimentation} data and the numbers reported in the original paper result from the fact that a different set of training data was used for training of the models.) 

\subsubsection{Panoptic segmentation:}
The results for the metrics related to the performance of panoptic and instance segmentation for both data sets are show in Tab.~\ref{tab:resultsInst}. 
The fully supervised approach without consideration of the \textit{instance decoder} (\textbf{Seg} variant) achieves values for PQ of 39.0\% and 32.3\% on the two data sets, and a $F_1^\text{inst}$ score of 51.1\% and 45.0\% respectively. 
\begin{table}[ht]
	\centering
	\caption{Results for the panoptic segmentation metrics achieved on the \textit{sedimentation} and \textit{fresh concrete} data sets.}
		\begin{tabular}{|l| c c | c c |} \hline
		&\multicolumn{2}{c|}{\textit{Sedimentation}} & \multicolumn{2}{c|}{\textit{Fresh concrete}}\\
		in [\%]& PQ & $F_1^\text{inst}$ & PQ & $F_1^\text{inst}$ \\\hline		
		Ours (\textbf{Seg})  							& 39.0 & 51.1 & 32.3 & 45.0  \\
		Ours (\textbf{Inst}) 							& 41.2 & 53.9 & 30.5 & 43.2  \\
		Ours (\textbf{ConsInst})  				& \textbf{48.2} & \textbf{61.9} & \textbf{39.0} & \textbf{52.6}\\ \hline
		Ouali et al. (2020) \cite{Ouali2020} 		& 43.9 & 57.7 & 34.5 & 47.5   \\ 
		Coenen et al. (2021) \cite{Coenen2021a}	& 47.6 & 61.8 & 33.7 & 46.6  \\ \hline
		\end{tabular}	
\label{tab:resultsInst}
\end{table}  
As is visible from Tab.~\ref{tab:resultsInst}, learning to predict the proposed instance representations from the limited amount of labelled data, in addition to the semantic segmentation, increases the results for PQ and $F_1^\text{inst}$ score by up to 2.8\% on the \textit{sedimentation data}. 
However, regarding the \textit{fresh concrete} data set, the PQ and $F_1^\text{inst}$ score achieved by the \textbf{Inst} variant are decreased by 1.8\% compared to the \textbf{Seg} setting. 
One assumption for the cause of the latter effect is, that due to the indistinctly defined instance boundaries in the images of that data set, a property that was already mentioned in Sec.~\ref{sec:datasets}, the network has to learn to guess parts of the object boundaries. 
Since the pixelwise instance representations proposed in this work are defined based on the location of each pixel relative to its closest boundary point, an implicit inference of the instance boundaries by the network is a prerequisite in order to predict the correct instance representing maps. 
We assume that the limited amount of labelled data used for training, together with the property of poorly visible and indiscernible instance boundaries in the images, causes the \textit{Inst} variant to perform worse on the \textit{fresh concrete} data.
Assumingly, the amount of labelled training data is not sufficient for the network to learn to infer the instance's extent when the boundaries are not clearly represented in the image data. 
However, as can be seen from the table, when introducing additional unlabelled data to the training process by applying our \textit{ConsInstancy} regularisation, we achieve the by far best results compared to the variants where no semi-supervised learning is applied. 
The semi-supervised variant (\textbf{ConsInst}) performs by up to 8.0\% better compared to the purely supervised variant (\textbf{Inst}) on the \textit{sedimentation} data and by up to 9.4\% better on the \textit{fresh concrete} data. 
Compared to the semi-supervised methods proposed in \cite{Coenen2021a, Ouali2020}, our approach performs slightly better on the \textit{sedimentation data} and outperforms the approaches by a large margin on our \textit{fresh concrete} data set.

\section{Conclusion}
We present a framework for semi-supervised panoptic segmentation based on the consensus principle.
To this end, we propose novel instance representations and a novel semi-supervised training scheme, denoted as \textit{ConsInstancy} training, by enforcing consistency during training between the multi-task predictions of the instance representations and a semantic segmentation map using entirely unlabelled data. Results on two data sets demonstrate the benefit of our multi-task framework using the proposed instance representations as well as the semi-supervised training on both tasks, semantic and panoptic segmentation of concrete aggregate particles.
A quantitative comparison showes that our approach is able to outperform current state-of-the-art methods for semi-supervised segmentation.
In the future, we aim at adapting and applying our framework on scenes with very dense instance occurrences, like e.g.\, piles of raw aggregate material, in which nearly every pixel belongs to an instances within the \textit{thing} classes.
Furthermore, we want to apply our framework on multi-class segmentation tasks by discerning between different aggregate types, as e.g.\,natural particles or recycled material, which can deliver valuable cues for requirements on the concrete composition and its mixture design.

\backmatter

\bmhead{Acknowledgments}
This work was supported by the Federal Ministry of Education and Research of Germany (BMBF) as part of the research project ReCyCONtrol [Project number 033R260A] (\url{https://www.recycontrol.uni-hannover.de}).

\bmhead{Statements and Declarations}
The authors have no competing interests to declare that are relevant to the content of this article.

\bibliographystyle{sn-mathphys}
\bibliography{Literatur}


\begin{thebibliography}{32}
\ifx \bisbn   \undefined \def \bisbn  #1{ISBN #1}\fi
\ifx \binits  \undefined \def \binits#1{#1}\fi
\ifx \bauthor  \undefined \def \bauthor#1{#1}\fi
\ifx \batitle  \undefined \def \batitle#1{#1}\fi
\ifx \bjtitle  \undefined \def \bjtitle#1{#1}\fi
\ifx \bvolume  \undefined \def \bvolume#1{\textbf{#1}}\fi
\ifx \byear  \undefined \def \byear#1{#1}\fi
\ifx \bissue  \undefined \def \bissue#1{#1}\fi
\ifx \bfpage  \undefined \def \bfpage#1{#1}\fi
\ifx \blpage  \undefined \def \blpage #1{#1}\fi
\ifx \burl  \undefined \def \burl#1{\textsf{#1}}\fi
\ifx \doiurl  \undefined \def \doiurl#1{\url{https://doi.org/#1}}\fi
\ifx \betal  \undefined \def \betal{\textit{et al.}}\fi
\ifx \binstitute  \undefined \def \binstitute#1{#1}\fi
\ifx \binstitutionaled  \undefined \def \binstitutionaled#1{#1}\fi
\ifx \bctitle  \undefined \def \bctitle#1{#1}\fi
\ifx \beditor  \undefined \def \beditor#1{#1}\fi
\ifx \bpublisher  \undefined \def \bpublisher#1{#1}\fi
\ifx \bbtitle  \undefined \def \bbtitle#1{#1}\fi
\ifx \bedition  \undefined \def \bedition#1{#1}\fi
\ifx \bseriesno  \undefined \def \bseriesno#1{#1}\fi
\ifx \blocation  \undefined \def \blocation#1{#1}\fi
\ifx \bsertitle  \undefined \def \bsertitle#1{#1}\fi
\ifx \bsnm \undefined \def \bsnm#1{#1}\fi
\ifx \bsuffix \undefined \def \bsuffix#1{#1}\fi
\ifx \bparticle \undefined \def \bparticle#1{#1}\fi
\ifx \barticle \undefined \def \barticle#1{#1}\fi
\bibcommenthead
\ifx \bconfdate \undefined \def \bconfdate #1{#1}\fi
\ifx \botherref \undefined \def \botherref #1{#1}\fi
\ifx \url \undefined \def \url#1{\textsf{#1}}\fi
\ifx \bchapter \undefined \def \bchapter#1{#1}\fi
\ifx \bbook \undefined \def \bbook#1{#1}\fi
\ifx \bcomment \undefined \def \bcomment#1{#1}\fi
\ifx \oauthor \undefined \def \oauthor#1{#1}\fi
\ifx \citeauthoryear \undefined \def \citeauthoryear#1{#1}\fi
\ifx \endbibitem  \undefined \def \endbibitem {}\fi
\ifx \bconflocation  \undefined \def \bconflocation#1{#1}\fi
\ifx \arxivurl  \undefined \def \arxivurl#1{\textsf{#1}}\fi
\csname PreBibitemsHook\endcsname

\bibitem{Kirillov2019}
\begin{bchapter}
\bauthor{\bsnm{Kirillov}, \binits{A.}},
\bauthor{\bsnm{He}, \binits{K.}},
\bauthor{\bsnm{Girshick}, \binits{R.}},
\bauthor{\bsnm{Rother}, \binits{C.}},
\bauthor{\bsnm{Doll{\'a}r}, \binits{P.}}:
\bctitle{{Panoptic Segmentation}}.
In: \bbtitle{IEEE Conference on Computer Vision and Pattern Recognition
  (CVPR)},
pp. \bfpage{9404}--\blpage{9413}
(\byear{2019})
\end{bchapter}
\endbibitem

\bibitem{Coenen2021a}
\begin{bchapter}
\bauthor{\bsnm{Coenen}, \binits{M.}},
\bauthor{\bsnm{Schack}, \binits{T.}},
\bauthor{\bsnm{Beyer}, \binits{D.}},
\bauthor{\bsnm{Heipke}, \binits{C.}},
\bauthor{\bsnm{Haist}, \binits{M.}}:
\bctitle{{Semi-Supervised Segmentation of Concrete Aggregate Using Consensus
  Regularisation and Prior Guidance}}.
In: \bbtitle{ISPRS Annals of Photogrammetry, Remote Sensing and Spatial
  Information Sciences},
vol. \bseriesno{V-2-2021},
pp. \bfpage{83}--\blpage{91}
(\byear{2021}).
\doiurl{10.5194/isprs-annals-V-2-2021-83-2021}
\end{bchapter}
\endbibitem

\bibitem{Wang2022}
\begin{botherref}
\oauthor{\bsnm{Wang}, \binits{W.}},
\oauthor{\bsnm{Su}, \binits{C.}},
\oauthor{\bsnm{Zhang}, \binits{H.}}:
{Automatic Segmentation of Concrete Aggregate using Convolutional Neural
  Network}.
Automation in Construction
\textbf{134}
(2022).
\doiurl{10.1016/j.autcon.2021.104106}
\end{botherref}
\endbibitem

\bibitem{Xiaomeng2021}
\begin{barticle}
\bauthor{\bsnm{Li}, \binits{X.}},
\bauthor{\bsnm{Yu}, \binits{L.}},
\bauthor{\bsnm{Chen}, \binits{H.}},
\bauthor{\bsnm{Fu}, \binits{C.-W.}},
\bauthor{\bsnm{Xing}, \binits{L.}},
\bauthor{\bsnm{Heng}, \binits{P.-A.}}:
\batitle{{Transformation-Consistent Self-Ensembling Model for Semisupervised
  Medical Image Segmentation}}.
\bjtitle{{IEEE Transactions on Neural Networks and Learning Systems}}
\bvolume{32}(\bissue{2}),
\bfpage{523}--\blpage{534}
(\byear{2021}).
\doiurl{10.1109/TNNLS.2020.2995319}
\end{barticle}
\endbibitem

\bibitem{Ouali2020}
\begin{bchapter}
\bauthor{\bsnm{Ouali}, \binits{Y.}},
\bauthor{\bsnm{Hudelot}, \binits{C.}},
\bauthor{\bsnm{Tami}, \binits{M.}}:
\bctitle{{Semi-Supervised Semantic Segmentation with Cross-Consistency
  Training}}.
In: \bbtitle{IEEE Conference on Computer Vision and Pattern Recognition
  (CVPR)},
pp. \bfpage{12674}--\blpage{12684}
(\byear{2020})
\end{bchapter}
\endbibitem

\bibitem{Peng2020}
\begin{botherref}
\oauthor{\bsnm{Peng}, \binits{J.}},
\oauthor{\bsnm{Estrada}, \binits{G.}},
\oauthor{\bsnm{Pedersoli}, \binits{M.}},
\oauthor{\bsnm{Desrosiers}, \binits{C.}}:
{Deep Co-Training for Semi-Supervised Image Segmentation}.
Pattern Recognition
\textbf{107}
(2020).
\doiurl{10.1016/j.patcog.2020.107269.}
\end{botherref}
\endbibitem

\bibitem{Zhang2020}
\begin{bchapter}
\bauthor{\bsnm{Zhang}, \binits{B.}},
\bauthor{\bsnm{Zhang}, \binits{Y.}},
\bauthor{\bsnm{Li}, \binits{Y.}},
\bauthor{\bsnm{Wan}, \binits{Y.}},
\bauthor{\bsnm{Wen}, \binits{F.}}:
\bctitle{{Semi-Supervised Semantic Segmentation Network via Learning
  Consistency for Remote Sensing Land-Cover Classification}}.
In: \bbtitle{ISPRS Annals of the Photogrammetry, Remote Sensing and Spatial
  Information Sciences},
vol. \bseriesno{V-2-2020},
pp. \bfpage{609}--\blpage{615}
(\byear{2020}).
\doiurl{10.5194/isprs-annals-V-2-2020-609-2020}
\end{bchapter}
\endbibitem

\bibitem{Qizhu2018}
\begin{bchapter}
\bauthor{\bsnm{Li}, \binits{Q.}},
\bauthor{\bsnm{Arnab}, \binits{A.}},
\bauthor{\bsnm{Torr}, \binits{P.}}:
\bctitle{{Weakly- and semi-supervised Panoptic Segmentation}}.
In: \bbtitle{European Conference on Computer Vision (ECCV)},
pp. \bfpage{102}--\blpage{118}
(\byear{2018}).
\doiurl{10.1007/978-3-030-01267-0\_}
\end{bchapter}
\endbibitem

\bibitem{Chao2016}
\begin{barticle}
\bauthor{\bsnm{Chao}, \binits{G.}},
\bauthor{\bsnm{Sun}, \binits{S.}}:
\batitle{{Consensus and Complementarity based maximum Entropy Discrimination
  for multi-view Classification}}.
\bjtitle{Information Sciences}
\bvolume{367-368},
\bfpage{296}--\blpage{310}
(\byear{2016}).
\doiurl{10.1016/j.ins.2016.06.004}
\end{barticle}
\endbibitem

\bibitem{FasterRCNN}
\begin{bchapter}
\bauthor{\bsnm{Ren}, \binits{S.}},
\bauthor{\bsnm{He}, \binits{K.}},
\bauthor{\bsnm{Girshick}, \binits{R.}},
\bauthor{\bsnm{Sun}, \binits{J.}}:
\bctitle{{Faster R-CNN: Towards Real-Time Object Detection with Region Proposal
  Networks}}.
In: \bbtitle{Advances in Neural Information Processing Systems (NIPS)},
vol. \bseriesno{28},
pp. \bfpage{91}--\blpage{99}
(\byear{2015})
\end{bchapter}
\endbibitem

\bibitem{MaskRCNN}
\begin{bchapter}
\bauthor{\bsnm{He}, \binits{K.}},
\bauthor{\bsnm{Gkioxari}, \binits{G.}},
\bauthor{\bsnm{Doll{\'a}r}, \binits{P.}},
\bauthor{\bsnm{Girshick}, \binits{R.}}:
\bctitle{{Mask R-CNN}}.
In: \bbtitle{IEEE International Conference on Computer Vision (ICCV)},
pp. \bfpage{2980}--\blpage{2988}
(\byear{2017}).
\doiurl{10.1109/ICCV.2017.322}
\end{bchapter}
\endbibitem

\bibitem{CentroidNet2019}
\begin{bchapter}
\bauthor{\bsnm{Dijkstra}, \binits{K.}},
\bauthor{\bparticle{van~de} \bsnm{Loosdrecht}, \binits{J.}},
\bauthor{\bsnm{Schomaker}, \binits{L.R.B.}},
\bauthor{\bsnm{Wiering}, \binits{M.A.}}:
\bctitle{{CentroidNet: A Deep Neural Network for Joint Object Localization and
  Counting}}.
In: \bbtitle{Machine Learning and Knowledge Discovery in Databases},
pp. \bfpage{585}--\blpage{601}
(\byear{2019}).
\doiurl{10.1007/978-3-030-10997-4_36}
\end{bchapter}
\endbibitem

\bibitem{CentroidNetV22021}
\begin{barticle}
\bauthor{\bsnm{Dijkstra}, \binits{K.}},
\bauthor{\bsnm{{van de Loosdrecht}}, \binits{J.}},
\bauthor{\bsnm{Atsma}, \binits{W.A.}},
\bauthor{\bsnm{Schomaker}, \binits{L.R.B.}},
\bauthor{\bsnm{Wiering}, \binits{M.A.}}:
\batitle{{CentroidNetV2: A hybrid Deep Neural Network for Small-Object
  Segmentation and Counting}}.
\bjtitle{Neurocomputing}
\bvolume{423},
\bfpage{490}--\blpage{505}
(\byear{2021}).
\doiurl{10.1016/j.neucom.2020.10.075}
\end{barticle}
\endbibitem

\bibitem{Xie2021}
\begin{barticle}
\bauthor{\bsnm{Xie}, \binits{E.}},
\bauthor{\bsnm{Wang}, \binits{W.}},
\bauthor{\bsnm{Ding}, \binits{M.}},
\bauthor{\bsnm{Zhang}, \binits{R.}},
\bauthor{\bsnm{Luo}, \binits{P.}}:
\batitle{{PolarMask++: Enhanced Polar Representation for Single-Shot Instance
  Segmentation and Beyond}}.
\bjtitle{IEEE Transactions on Pattern Analysis and Machine Intelligence
  (TPAMI)}
(\byear{2021}).
\doiurl{10.1109/TPAMI.2021.3080324}
\end{barticle}
\endbibitem

\bibitem{Schmidt2018}
\begin{bchapter}
\bauthor{\bsnm{Schmidt}, \binits{U.}},
\bauthor{\bsnm{Weigert}, \binits{M.}},
\bauthor{\bsnm{Broaddus}, \binits{C.}},
\bauthor{\bsnm{Myers}, \binits{G.}}:
\bctitle{{Cell Detection with Star-Convex Polygons}}.
In: \bbtitle{International Conference on Medical Image Computing and
  Computer-Assisted Intervention (MICCAI)},
pp. \bfpage{265}--\blpage{273}
(\byear{2018}).
\doiurl{10.1007/978-3-030-00934-2_30}
\end{bchapter}
\endbibitem

\bibitem{Weigert2020}
\begin{bchapter}
\bauthor{\bsnm{Weigert}, \binits{M.}},
\bauthor{\bsnm{Schmidt}, \binits{U.}},
\bauthor{\bsnm{Haase}, \binits{R.}},
\bauthor{\bsnm{Sugawara}, \binits{K.}},
\bauthor{\bsnm{Myers}, \binits{G.}}:
\bctitle{{Star-Convex Polyhedra for 3D Object Detection and Segmentation in
  Microscopy}}.
In: \bbtitle{IEEE Winter Conference on Applications of Computer Vision (WACV)},
pp. \bfpage{3666}--\blpage{3673}
(\byear{2020}).
\doiurl{10.1109/WACV45572.2020.9093435}
\end{bchapter}
\endbibitem

\bibitem{Bai2017}
\begin{bchapter}
\bauthor{\bsnm{Bai}, \binits{M.}},
\bauthor{\bsnm{Urtasun}, \binits{R.}}:
\bctitle{{Deep Watershed Transform for Instance Segmentation}}.
In: \bbtitle{IEEE Conference on Computer Vision and Pattern Recognition
  (CVPR)},
pp. \bfpage{2858}--\blpage{2866}
(\byear{2017}).
\doiurl{10.1109/CVPR.2017.305}
\end{bchapter}
\endbibitem

\bibitem{Kalluri2019}
\begin{bchapter}
\bauthor{\bsnm{Kalluri}, \binits{T.}},
\bauthor{\bsnm{Varma}, \binits{G.}},
\bauthor{\bsnm{Chandraker}, \binits{M.}},
\bauthor{\bsnm{Jawahar}, \binits{C.V.}}:
\bctitle{{Universal Semi-Supervised Semantic Segmentation}}.
In: \bbtitle{IEEE International Conference on Computer Vision (ICCV)},
pp. \bfpage{5259}--\blpage{5270}
(\byear{2019})
\end{bchapter}
\endbibitem

\bibitem{Wittich2020}
\begin{bchapter}
\bauthor{\bsnm{Wittich}, \binits{D.}}:
\bctitle{{Deep Domain Adaptation by weighted Entropy Minimization for the
  Classification of Aerial Images}}.
In: \bbtitle{ISPRS Annals of Photogrammetry, Remote Sensing and Spatial
  Information Sciences},
vol. \bseriesno{V-2-2020},
pp. \bfpage{591}--\blpage{598}
(\byear{2020}).
\doiurl{10.5194/isprs-annals-V-2-2020-591-2020}
\end{bchapter}
\endbibitem

\bibitem{Luc2016}
\begin{bchapter}
\bauthor{\bsnm{Luc}, \binits{P.}},
\bauthor{\bsnm{Couprie}, \binits{C.}},
\bauthor{\bsnm{Chintala}, \binits{S.}},
\bauthor{\bsnm{Verbeek}, \binits{J.}}:
\bctitle{{Semantic Segmentation using Adversarial Networks}}.
In: \bbtitle{NIPS Workshop on Adversarial Training}
(\byear{2016})
\end{bchapter}
\endbibitem

\bibitem{Souly2017}
\begin{bchapter}
\bauthor{\bsnm{Souly}, \binits{N.}},
\bauthor{\bsnm{Spampinato}, \binits{C.}},
\bauthor{\bsnm{Shah}, \binits{M.}}:
\bctitle{{Semi Supervised Semantic Segmentation Using Generative Adversarial
  Network}}.
In: \bbtitle{IEEE International Conference on Computer Vision (ICCV)},
pp. \bfpage{5689}--\blpage{5697}
(\byear{2017}).
\doiurl{10.1109/ICCV.2017.606}
\end{bchapter}
\endbibitem

\bibitem{Hung2018}
\begin{bchapter}
\bauthor{\bsnm{Hung}, \binits{W.C.}},
\bauthor{\bsnm{Tsai}, \binits{Y.H.}},
\bauthor{\bsnm{Liou}, \binits{Y.T.}},
\bauthor{\bsnm{Lin}, \binits{Y.Y.}},
\bauthor{\bsnm{Yang}, \binits{M.H.}}:
\bctitle{{Adversarial Learning forSemi-Supervised Semantic Segmentation}}.
In: \bbtitle{British Machine Vision Conference (BMVC)}
(\byear{2018})
\end{bchapter}
\endbibitem

\bibitem{Mendel2020}
\begin{bchapter}
\bauthor{\bsnm{Mendel}, \binits{R.}},
\bauthor{\bparticle{de} \bsnm{Souza}, \binits{L.A.}},
\bauthor{\bsnm{Rauber}, \binits{D.}},
\bauthor{\bsnm{Papa}, \binits{J.P.}},
\bauthor{\bsnm{Palm}, \binits{C.}}:
\bctitle{{Semi-supervised Segmentation Based on Error-Correcting Supervision}}.
In: \bbtitle{European Conference on Computer Vision (ECCV)},
pp. \bfpage{141}--\blpage{157}
(\byear{2020}).
\doiurl{10.1007/978-3-030-58526-6_9}
\end{bchapter}
\endbibitem

\bibitem{Myronenko2019}
\begin{bchapter}
\bauthor{\bsnm{Myronenko}, \binits{A.}}:
\bctitle{{3D MRI Brain Tumor Segmentation Using Autoencoder Regularization}}.
In: \bbtitle{International MICCAI Brainlesion Workshop}.
\bsertitle{Lecture Notes in Computer Science},
vol. \bseriesno{11384},
pp. \bfpage{311}--\blpage{320}
(\byear{2019}).
\doiurl{10.1007/978-3-030-11726-9_28}
\end{bchapter}
\endbibitem

\bibitem{Sedai2017}
\begin{bchapter}
\bauthor{\bsnm{Sedai}, \binits{S.}},
\bauthor{\bsnm{Mahapatra}, \binits{D.}},
\bauthor{\bsnm{Hewavitharanage}, \binits{S.}},
\bauthor{\bsnm{Maetschke}, \binits{S.}},
\bauthor{\bsnm{Garnavi}, \binits{R.}}:
\bctitle{{Semi-Supervised Segmentation of Optic Cup in Retinal Fundus Images
  Using Variational Autoencoder}}.
In: \bbtitle{Medical Image Computing and Computer-Assisted Intervention
  (MICCAI)}.
\bsertitle{Lecture Notes in Computer Science},
vol. \bseriesno{10434},
pp. \bfpage{75}--\blpage{82}
(\byear{2017}).
\doiurl{10.1007/978-3-319-66185-8_9}
\end{bchapter}
\endbibitem

\bibitem{Xiaomeng2018}
\begin{bchapter}
\bauthor{\bsnm{Li}, \binits{X.}},
\bauthor{\bsnm{Yu}, \binits{L.}},
\bauthor{\bsnm{Chen}, \binits{H.}},
\bauthor{\bsnm{Fu}, \binits{C.W.}},
\bauthor{\bsnm{Heng}, \binits{P.A.}}:
\bctitle{{Semi-supervised Skin Lesion Segmentation via Transformation
  Consistent Self-ensembling Model}}.
In: \bbtitle{British Machine Vision Conference (BMVC)}
(\byear{2018})
\end{bchapter}
\endbibitem

\bibitem{Hao2021}
\begin{botherref}
\oauthor{\bsnm{Hao}, \binits{S.}},
\oauthor{\bsnm{Wang}, \binits{G.}},
\oauthor{\bsnm{Gu}, \binits{R.}}:
{Weakly Supervised Instance Segmentation using Multi-Prior Fusion}.
Computer Vision and Image Understanding
\textbf{211}
(2021).
\doiurl{10.1016/j.cviu.2021.103261}
\end{botherref}
\endbibitem

\bibitem{Hsu2019}
\begin{bchapter}
\bauthor{\bsnm{Hsu}, \binits{C.-C.}},
\bauthor{\bsnm{Hsu}, \binits{K.-J.}},
\bauthor{\bsnm{Tsai}, \binits{C.-C.}},
\bauthor{\bsnm{Lin}, \binits{Y.-Y.}},
\bauthor{\bsnm{Chuang}, \binits{Y.-Y.}}:
\bctitle{{Weakly Supervised Instance Segmentation using the Bounding Box
  Tightness Prior}}.
In: \bbtitle{Advances in Neural Information Processing Systems},
vol. \bseriesno{32},
pp. \bfpage{6586}--\blpage{6597}
(\byear{2019})
\end{bchapter}
\endbibitem

\bibitem{Felzenswalb2012}
\begin{barticle}
\bauthor{\bsnm{Felzenszwalb}, \binits{P.F.}},
\bauthor{\bsnm{Huttenlocher}, \binits{D.P.}}:
\batitle{{Distance Transforms of sampled Functions}}.
\bjtitle{Theory of Computing}
\bvolume{8}(\bissue{1}),
\bfpage{415}--\blpage{428}
(\byear{2012})
\end{barticle}
\endbibitem

\bibitem{Gambhir2004}
\begin{bbook}
\bauthor{\bsnm{Gambhir}, \binits{M.L.}}:
\bbtitle{Concrete Technology}.
\bsertitle{Civil engineering series},
(\byear{2004})
\end{bbook}
\endbibitem

\bibitem{He2015}
\begin{bchapter}
\bauthor{\bsnm{He}, \binits{K.}},
\bauthor{\bsnm{Zhang}, \binits{X.}},
\bauthor{\bsnm{Ren}, \binits{S.}},
\bauthor{\bsnm{Sun}, \binits{J.}}:
\bctitle{{Delving Deep into Rectifiers: Surpassing Human-Level Performance on
  ImageNet Classification}}.
In: \bbtitle{IEEE International Conference on Computer Vision (ICCV)},
pp. \bfpage{1026}--\blpage{1034}
(\byear{2015}).
\doiurl{10.1109/ICCV.2015.123}
\end{bchapter}
\endbibitem

\bibitem{Adam2015}
\begin{bchapter}
\bauthor{\bsnm{Kingma}, \binits{D.P.}},
\bauthor{\bsnm{Ba}, \binits{L.J.}}:
\bctitle{{Adam: A Method for Stochastic Optimization}}.
In: \bbtitle{International Conference on Learning Representations (ICLR)}
(\byear{2015})
\end{bchapter}
\endbibitem

\end{thebibliography}


\end{document}